\documentclass[11pt]{article}

\usepackage[final]{acl}

\usepackage{times}
\usepackage{latexsym}
\usepackage{graphicx}
\usepackage{booktabs}
\usepackage{xcolor}
\usepackage{xspace}
\usepackage{multirow}
\usepackage{enumitem}
\usepackage[T1]{fontenc}

\usepackage[utf8]{inputenc}

\usepackage{microtype}

\usepackage{inconsolata}
\usepackage{pgfplots}
\usepgfplotslibrary{groupplots}
\usepackage{placeins}
\pgfplotsset{compat=1.18}

\usepackage{graphicx}
\newcommand{\ours}{\textsc{CrossAug}\xspace}
\title{Beyond Chunk-Local Extraction: Cross-Chunk Graph Augmentation for GraphRAG}

\author{
\textbf{Jiaming Zhang\textsuperscript{1}},
\textbf{Yibo Zhao\textsuperscript{1}},
\textbf{Jing Yu\textsuperscript{1}},
\\
\textbf{Jianxiang Yu\textsuperscript{1}},
\textbf{Xiang Li\textsuperscript{1}\thanks{Corresponding author: \texttt{xiangli@dase.ecnu.edu.cn}}}
\\
\textsuperscript{1}School of Data Science and Engineering, East China Normal University
}

\begin{document}
\maketitle
\begin{abstract}
GraphRAG extends retrieval-augmented generation by organizing corpora as explicit knowledge graphs, enabling graph-based retrieval for complex question answering. However, existing frameworks extract entities and relations within individual chunks, leaving cross-chunk relations---those whose evidence spans multiple passages---systematically absent from the index. Exhaustive LLM-based recovery of such relations is impractical due to the combinatorial explosion of chunk combinations. We present \ours, a GNN-guided \underline{CROSS}-Chunk Graph \underline{AUG}mentation method that enriches GraphRAG indices with cross-chunk relational structure as an offline step before query-time retrieval. \ours derives training supervision through self-supervised graph corruption, uses a topology-aware GNN to score subgraphs for missingness, and applies evidence-grounded LLM completion only to selected high-scoring regions. Experiments on three LLM-based GraphRAG frameworks across four multi-hop and long-document QA benchmarks demonstrate that \ours consistently improves performance, confirming the benefit of cross-chunk graph augmentation for retrieval-based question answering. Our code is available at \url{https://github.com/DonFinliani/CrossAug}.

\end{abstract}

\section{Introduction}

Retrieval-augmented generation (RAG) is widely used in processing long documents, where documents are typically chunked into smaller segments for retrieval and generation. Recent GraphRAG frameworks extend this paradigm by organizing the corpus as an explicit graph in which entities are extracted from text and linked via relation triples, enabling graph algorithms and community summarization to support multi-hop retrieval and corpus-level queries \citep{edge2024local, gutierrez2024hipporag}. This direction has shown particular promise for long texts and complex QA, where flat vector retrieval frequently surfaces superficially relevant information while missing the critical evidence required for complex question answering.

{However, the conventional chunk-based construction of GraphRAG introduces a fundamental blind spot. Existing frameworks typically extract entities and relations within individual chunks, implicitly assuming that document-level relational structure can be recovered by stitching together locally extracted triples. This assumption is fragile for long texts, where many important relations, such as alias resolution and causal chains, depend on evidence distributed across multiple chunks. For example, when HippoRAG2~\citep{gutierrez2025rag} constructs a graph index for \textit{Voodoo Planet}, the resulting graph contains locally correct triples such as \textit{(Tau, approached, Lumbrilo)} and \textit{(Tau, defeated, Lumbrilo)}, but these can only form an event chain around Tau and Lumbrilo without specifying where the confrontation takes place, causing the framework to answer incorrectly. The cross-chunk locative relation \textit{(Khatkan swamp, location of, final confrontation with Lumbrilo)} can only be recovered by jointly reading the chunk that describes the swamp with those describing the battle, which a chunk-local extractor has no access to. Community summarization solves corpus-level queries but cannot recover such precise cross-chunk relational facts and can even discard relational details rather than preserving them~\citep{han2025rag}. A GraphRAG index may thus contain many locally correct edges while still missing the hidden cross-chunk relations required for long-document reasoning.}

{Most existing GraphRAG methods focus on improving how an already constructed graph is used for retrieval, ranking, summarization, or multi-hop reasoning; if the graph is incomplete at construction time, however, downstream algorithms can only operate over a partial structure~\citep{yang2026graph}. While LLMs could in principle recover missing cross-chunk relations, applying them exhaustively is impractical due to the combinatorial explosion of chunk combinations. We therefore introduce graph neural networks (GNNs) into the GraphRAG pipeline as a structural guide: by combining semantic node features with relational graph structure through edge-type-aware message passing \citep{gilmer2017neural,schlichtkrull2018modeling}, a GNN can identify structural patterns characteristic of chunk-local extraction failures and locate where evidence-grounded LLM completion is most likely to recover useful relational structure.}

In this paper, we introduce \ours, a GNN-guided \underline{CROSS}-Chunk Graph \underline{AUG}mentation method for GraphRAG. \ours is built around three design principles. First, a \textit{self-supervised graph corruption module} creates training data by corrupting the base graph itself---masking fact edges and deleting entity nodes---to match the incompleteness patterns introduced by chunk-local extraction. Second, a \textit{topology-aware GNN scoring module} learns to assign missingness scores to local subgraphs, turning cross-chunk discovery from exhaustive entity enumeration into a targeted extraction problem with a budget hyperparameter. Third, a \textit{GNN-guided LLM completion module} applies the LLM only to selected, evidence-grounded subgraphs and persists validated extracted triples back into the graph index. In this way, the additional reasoning cost is paid once during offline augmentation, while the enriched relational structure can be reused by any downstream retrieval pipeline.

Finally, our contributions are threefold:
\begin{itemize}[noitemsep,topsep=2pt]
    \item We identify GraphRAG's chunk-local extraction bottleneck and formulate it as cross-chunk relation/entity recovery.
    \item We propose a self-supervised GNN scoring framework that reframes cross-chunk discovery as subgraph-level missingness detection, replacing intractable enumeration with targeted and topology-guided selection.
    \item We present a GNN--LLM collaboration pipeline that yields consistent performance improvements across three GraphRAG frameworks and four QA benchmarks.
\end{itemize}

\begin{figure*}[t]
\centering
\includegraphics[width=\textwidth]{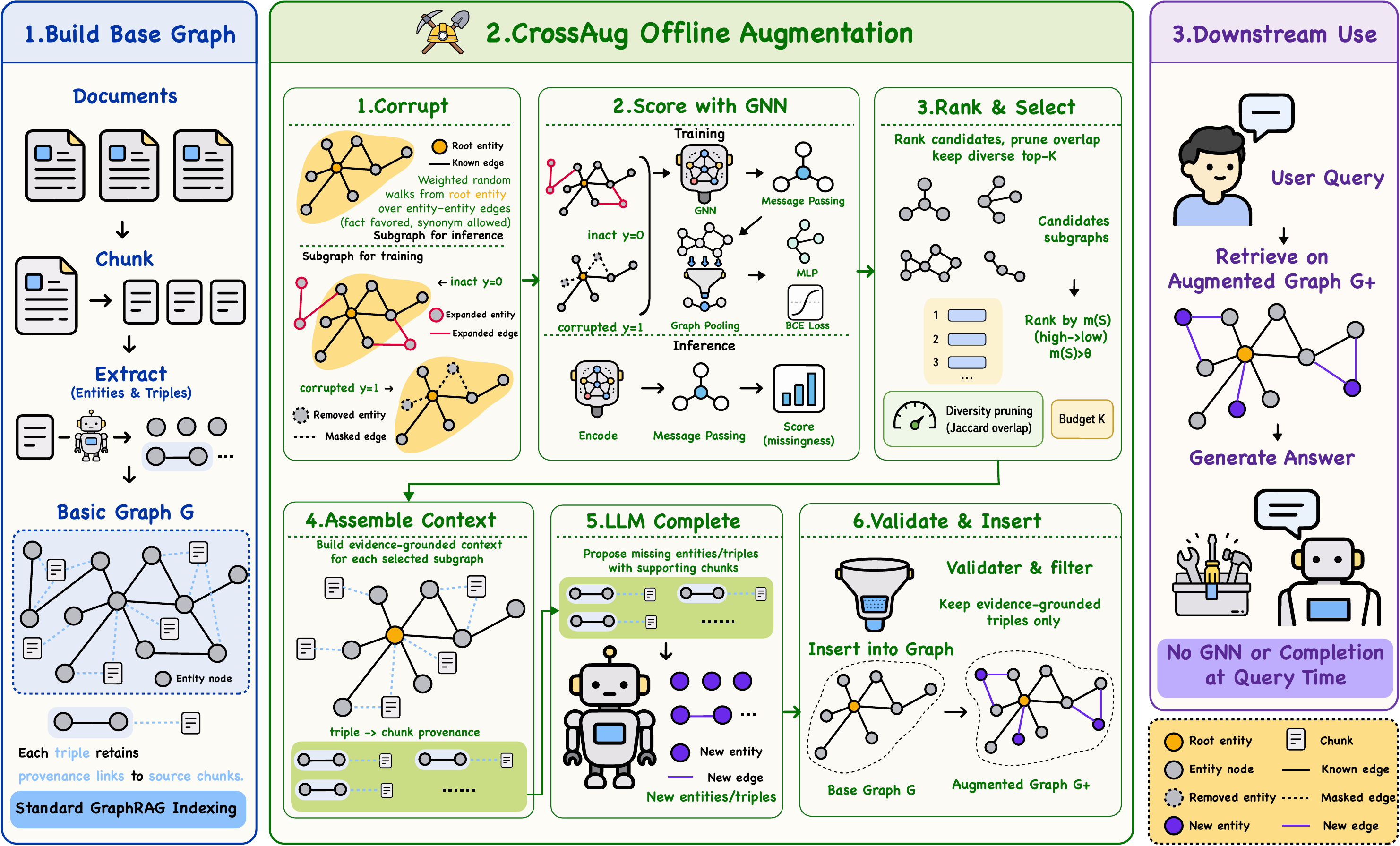}
\caption{Overview of the \ours workflow. GNN-based missingness scoring selects incomplete subgraphs, and LLM completion extracts evidence-grounded triples before augmenting the graph index.}
\label{fig:workflow}
\end{figure*}

\section{Related Work}

GraphRAG has recently gained increasing attention in the research community. MS-GraphRAG~\citep{edge2024local} is among the first GraphRAG frameworks to build a knowledge graph index directly from raw text for corpus-level question answering. Its pipeline demonstrates strong improvements over flat vector retrieval for broad queries, though graph construction requires substantial LLM inference.

Recent GraphRAG frameworks can be broadly grouped by how they construct the indexing graph. One line of work uses LLMs or OpenIE-style extraction to build semantic graphs with explicit entities and relations. HippoRAG and HippoRAG2 extract entities and (subject, predicate, object) triples from each chunk to build schema-free knowledge graphs, and then apply Personalized PageRank over entity, fact, and passage nodes to support associative multi-hop retrieval \citep{gutierrez2024hipporag,gutierrez2025rag}. LightRAG constructs a graph-enhanced index with entity and relation descriptions and combines low-level and high-level retrieval to improve both local factual access and broader contextual retrieval \citep{guo2024lightrag}. GFM-RAG extends this line by training a general GNN over the extracted graph to enable topology-aware retrieval beyond similarity-based ranking \citep{luo2026gfm}. A second line of work reduces indexing cost by avoiding heavy LLM relation extraction. For example, LinearRAG builds a relation-free hierarchical index with lightweight entity extraction and semantic linking, avoiding explicit relation-triple generation during graph construction \citep{zhuang2025linearrag}. E$^2$GraphRAG similarly focuses on efficient index construction with NLP-tool-based entity graphs and bidirectional entity--chunk links, with summaries used only as lightweight retrieval support \citep{zhao20252graphrag}. These lightweight frameworks trade explicit semantic relation modeling for lower construction cost and higher scalability. Despite their different cost--quality trade-offs, both families construct graphs primarily from chunk-local evidence: LLM-based frameworks usually extract triples within individual chunks, while lightweight frameworks rely on local entity mentions, co-occurrence, or chunk-level indexes. As a result, neither family directly addresses relations that can only be recovered by jointly reading evidence distributed across multiple chunks.

\section{Method}

\subsection{Overview}

{Given a long document collection, a standard GraphRAG pipeline splits the text into chunks, extracts entities and triples, and builds a graph index $\mathcal{G}=(\mathcal{V},\mathcal{E})$ for retrieval. Nodes include entity, passage, and auxiliary nodes; edges include fact edges from extracted triples, synonym edges, and entity--chunk edges. Although this graph supports graph-based retrieval, it remains incomplete because chunk-local extraction can miss both entities and fact edges. The goal of \ours is to discover missing entities and hidden triples that are supported by the original text but absent from the graph, especially when their evidence is distributed across multiple chunks.}

\ours is applied as an offline augmentation step after the base graph has been constructed by the GraphRAG pipeline. It does not modify the query-time retrieval or generation components; instead, it enriches the graph index once before any downstream use. \ours operates through a three-module pipeline. First, the \textit{self-supervised graph corruption module} builds training data from the existing graph itself. It samples local entity-centered subgraphs and constructs incomplete views by masking fact edges or deleting entity nodes. This converts the absence of manually labeled missing triples into a self-supervised missingness detection task. Second, the \textit{topology-aware GNN scoring module} learns to assign a missingness score to a sampled subgraph. The GNN does not predict a particular missing edge; instead, it estimates whether the local topology looks incomplete, which is better aligned with the goal of deciding where the LLM should spend effort. Third, the \textit{GNN-guided LLM completion module} sends only high-scoring and low-overlap subgraphs to the LLM together with evidence chunks and known triples. The LLM produces new evidence-grounded triples, and the validated triples are inserted back into the GraphRAG index. In this way, \ours separates structural search from semantic completion: the GNN provides a cheap global filter over many candidate regions, while the LLM performs expensive extraction only on selected contexts.

\subsection{Self-Supervised Graph Corruption}

\paragraph{Heterogeneous graph construction.}
\ours is designed for the common family of GraphRAG frameworks that construct a graph from LLM-extracted triples. Such frameworks usually produce an entity-centric graph: entities are represented as nodes, extracted triples induce semantic relation edges between entities, and each relation is associated with its source chunk as textual provenance. \ours therefore assumes a minimal graph interface rather than a framework-specific schema: the base index should provide entity nodes, relation-bearing edges, and a way to trace extracted facts back to evidence chunks. Each node $v \in \mathcal{V}$ is assigned a semantic feature $x_v$ from the existing retrieval index when available. When the base graph stores edge weights, \ours uses the raw weight $w_e$ as an edge-strength signal, such as the number of extracted triples supporting a fact edge or the similarity score of an alias/synonym edge; otherwise, unweighted fact edges are assigned a default weight of $1$. Each edge is also assigned a relation type $\psi(e)$ that distinguishes semantic fact edges from auxiliary structural edges. This abstraction makes the method applicable to LLM-based GraphRAG indices beyond a specific GraphRAG implementation.

\paragraph{Weighted random-walk sampling.}
Since the full graph can be large, \ours does not enumerate all entity pairs or chunk pairs. Instead, it samples local entity subgraphs around root entities. A root entity is eligible if it is an entity node, {is not filtered out for its uninformative content (e.g., purely numeric strings)}, and has at least one neighboring entity through a fact or synonym edge. From each root $r$, \ours launches multiple weighted random walks over {entity--entity} edges. Let $e=(u,v)$ be a fact or synonym edge with raw graph weight $w_e$. The transition weight used for sampling is:
\[
\tilde{w}_e = \min(w_e,c)\cdot \alpha_{\psi(e)},
\]
where $c$ clips very large edge weights and $\alpha_{\psi(e)}$ is an edge-type multiplier. In our implementation, fact edges receive a larger multiplier than synonym edges, so walks prefer semantically informative factual connections while still allowing alias or synonym expansion. The transition probability is then calculated by:
\[
P(v\mid u)=\tilde{w}_{(u,v)} / \sum_{z\in N(u)}\tilde{w}_{(u,z)}.
\]
The walk length is measured in entity hops, and chunk nodes are not traversed during sampling. Evidence chunks are collected later from the provenance of fact edges. This choice keeps the sampled region focused on entity-level relational structure and prevents early expansion into broad passage neighborhoods.

The sampled node set defines a local subgraph $S_r=(V_r,E_r)$. \ours retains only subgraphs with at least a minimum number of nodes and at least one fact edge. This filtering removes degenerate samples with little training signal and ensures that each retained subgraph has sufficient relational structure for corruption and LLM completion.
\paragraph{Corruption as missingness supervision.}
Manually annotating which triples are missing from a long-document graph is expensive. 
Our method therefore derives training supervision through self-supervised learning. 
Note that at both training  and inference time, 
a candidate subgraph sampled from a root entity is fed directly to the GNN for missingness scoring. 
During training, the sampled subgraph serves as the base for constructing corrupted positive examples ($y=1$) by applying two corruption operators (details will be described below). 
The intact negative examples ($y=0$), however, are obtained by more extensively exploring the local neighborhood around the same root entity, yielding comparatively richer and larger subgraphs than those sampled at inference time. 
This design ensures that the GNN learns to distinguish structurally incomplete regions, 
rather than acquiring an incidental association between subgraph density and the missingness label.

Two corruption operators are used. The first operator masks fact edges. Given the fact-edge set $E_r^{\mathrm{fact}}$, \ours removes a random subset $M_e \subset E_r^{\mathrm{fact}}$ with ratio $\rho_e$, while preserving at least the required minimum number of fact edges. This simulates the common failure mode where an extractor recognizes the relevant entities but misses one or more relations. The second operator deletes entity nodes. It selects eligible non-root entity nodes that participate in at least one fact edge and removes a subset $M_v$ with ratio $\rho_v$, provided that the remaining graph still contains enough fact edges. This simulates a stronger failure mode where the graph misses an entity along with its related relations.

These two corruptions correspond to the two main types of incompleteness that appear in GraphRAG: missing relation edges and missing entity nodes. The corruption task is intentionally defined at the subgraph level rather than the edge level. A hidden cross-chunk fact such as alias or causal chain requires region-level relational context rather than a single isolated entity pair. By asking the model to detect whether a local region is incomplete, \ours learns a retrieval-control signal: which graph regions deserve LLM completion.

\subsection{Topology-Aware GNN Missingness Scoring}

\paragraph{GNN encoder.}
For every sampled subgraph view, \ours constructs a heterogeneous graph representation for message passing with node features, typed edges, and node-type identifiers. Let $x_v$ denote the input semantic feature vector of node $v$, and let $W_x$ be a learned projection matrix that maps input features into the GNN hidden space. In \ours, the textual content of entity nodes and chunk nodes is encoded by the same embedding model. Therefore, although the sampled graph contains different node types, their features lie in the same vector space and have the same dimensionality, allowing them to share the same projection matrix $W_x$. The missingness detector initializes each node representation by combining this projected feature with a learned embedding of the node type:
\[
h_v^{(0)} = W_x x_v + e_{\phi(v)}.
\]
The node representations are then updated through two layers of message passing. For relational GNNs, the update uses edge-type-specific transformations:
\[
h_v^{(\ell+1)} =
\sigma\!\left(
\sum_{\tau}\sum_{u\in N_{\tau}(v)}
W_{\tau}^{(\ell)} h_u^{(\ell)}
\right),
\]
where $\ell$ denotes the message-passing layer, $\tau$ ranges over edge types, $N_{\tau}(v)$ denotes the neighbors of node $v$ connected by edges of type $\tau$, $u$ is a neighboring node in this typed neighborhood, $W_{\tau}^{(\ell)}$ is the layer-specific transformation matrix for edge type $\tau$, and $\sigma(\cdot)$ is a nonlinear activation function.

\paragraph{Graph-level missingness prediction.}
{After message passing, \ours aggregates the node representations into a single subgraph representation. Since graph nodes have no fixed order, we use order-invariant pooling operations. Following common graph-level pooling practice, we concatenate global mean pooling and global max pooling \citep{wang2024graph}:
\[
g(S)=\left[\mathrm{mean}_{v\in S} h_v;\ \mathrm{max}_{v\in S} h_v\right].
\]
Mean pooling summarizes the overall distribution of node representations in the sampled subgraph, while max pooling preserves the strongest local activation. Their concatenation therefore captures both the global subgraph context and localized evidence of incompleteness.}

{A multilayer classifier maps $g(S)$ to a logit $a(S)$, and the missingness score is $m(S)=\sigma(a(S))$, where $\sigma(\cdot)$ denotes the sigmoid function.}The model is trained with binary cross-entropy. Intact sampled views are treated as negative examples and corrupted views as positive examples. 
\[
\mathcal{L}_{\mathrm{BCE}} =
-y\log m(S) - (1-y)\log(1-m(S)).
\]

\paragraph{Dynamic training and inference selection.}
At each training epoch, \ours re-samples a fixed number of root entities and constructs fresh intact/corrupted training groups. This dynamic sampling exposes the GNN to different local neighborhoods and different corruption patterns, reducing overfitting to a static set of artificial examples.

At inference time, the trained GNN scores each sampled subgraph, and candidates are ranked by $m(S)$. A subgraph is eligible for LLM completion only if its score exceeds a threshold $\theta$. Candidates with high node-level overlap with already-selected subgraphs---measured by Jaccard similarity $J(S_i,S_j)=|V_i\cap V_j|/|V_i\cup V_j|$---are skipped to avoid redundant LLM calls on nearly identical evidence. Selection continues until the LLM call budget $B$ is exhausted. The GNN module thereby transforms the problem from exhaustive enumeration of all possible cross-chunk combinations into targeted selection of a diverse, high-missingness subgraph set, improving cost efficiency by directing LLM completion toward contexts where it is most likely to recover useful relational information.

\subsection{GNN-Guided LLM Completion and Graph Augmentation}

For each selected subgraph, \ours assembles a context that includes the subgraph's root entity, the full list of entities in the local region, all already extracted triples, and the evidence passages retrieved from the provenance of participating fact edges. The GNN is used to identify regions that are likely to be incomplete, while the LLM performs completion based on the actual entities, triples, and evidence passages in those regions.

The LLM is instructed to propose new entities and relations that are directly stated or unambiguously supported by the provided passages. Every proposed entity and relation must include a list of supporting chunk identifiers drawn from the evidence set; this citation requirement enforces cross-chunk grounding and prevents hallucinated graph expansion. Relations without valid chunk citations are discarded. The validated triples are then inserted into the base graph index using the same schema as the original extractor, so the augmented graph is structurally consistent with the base and immediately compatible with any downstream retrieval pipeline. Because all augmentation is performed offline, the enriched graph serves as a static index for subsequent retrieval and question answering; neither the LLM completion step nor the GNN need to be re-invoked at query time.

\section{Experiments}
Due to space limitations, we defer additional results to the Appendix, including Cost and Efficiency Analysis (Appendix~\ref{sec:cost-sensitivity}) and Hyperparameter Sensitivity Analysis (Appendix~\ref{sec:param-sensitivity}).
\subsection{Experimental Settings}

\paragraph{Datasets and Baselines}
We evaluate our method on four benchmarks: three widely used multi-hop QA benchmarks---MusiQue~\citep{trivedi2021musique}, 2WikiMultiHopQA~\citep{ho2020constructing}, and HotpotQA~\citep{yang2018hotpotqa}---and LiteraryQA~\citep{bonomo2025literaryqa}, a long-document QA benchmark over literary texts. Following previous work~\citep{gutierrez2024hipporag}, we sample 1,000 questions from each multi-hop benchmark to control experimental cost. For LiteraryQA, we use the first 50 books from the test set, comprising 1,358 queries in total; most books in this set exceed the context window of the base model. Detailed dataset statistics are shown in Table~\ref{tab:datasets}.
\begin{table}[!htbp]
\centering
\small
\setlength{\tabcolsep}{2pt}
\renewcommand{\arraystretch}{1.15}
\begin{tabular}{@{}lrrrr@{}}
\hline
\textbf{Statistic} & \textbf{LiteraryQA} & \textbf{MuSiQue} & \textbf{2Wiki} & \textbf{HotpotQA} \\
\hline
Chunks & 9,550 & 11,656 & 6,119 & 9,811 \\
Base nodes & 68,014 & 104,190 & 54,487 & 84,632 \\
Base edges & 225,908 & 331,233 & 169,455 & 262,256 \\
Base triples & 103,148 & 123,308 & 64,523 & 98,592 \\
Added nodes & 3,203 & 2,994 & 1,569 & 1,718 \\
Added edges & 27,284 & 20,093 & 9,781 & 9,839 \\
Added triples & 8,707 & 8,901 & 5,048 & 5,252 \\
\hline
\end{tabular}
\caption{Dataset-level graph statistics for the base GraphRAG index and the \ours-augmented index. ``2Wiki'' denotes 2WikiMultiHopQA and ``HotpotQA'' denotes HotpotQA.}
\label{tab:datasets}
\end{table}
\begin{table*}[t]
\centering
\begingroup
\scriptsize
\setlength{\tabcolsep}{3pt}
\resizebox{0.75\textwidth}{!}{%
\begin{tabular}{@{}c l cccccccc@{}}
\toprule
\textbf{Framework} & \textbf{Setting} &
\multicolumn{2}{c}{\textbf{MuSiQue}} &
\multicolumn{2}{c}{\textbf{2Wiki}} &
\multicolumn{2}{c}{\textbf{HotpotQA}} &
\multicolumn{2}{c}{\textbf{LiteraryQA}} \\
\cmidrule(lr){3-4}
\cmidrule(lr){5-6}
\cmidrule(lr){7-8}
\cmidrule(lr){9-10}
& & \textbf{EM} & \textbf{F1} &
\textbf{EM} & \textbf{F1} &
\textbf{EM} & \textbf{F1} &
\textbf{EM} & \textbf{F1} \\
\midrule

LinearRAG & --
& 24.10 & 37.23
& 47.90 & \textbf{65.16}
& 47.00 & 64.59
& 11.56 & 26.57 \\
\midrule

\multirow{2}{*}{LightRAG} & Base
& 20.50 & 28.09
& 28.40 & 36.93
& 44.90 & 58.60
& 11.26 & 26.24 \\

& \ours
& 21.40 & 29.24
& 29.30 & 37.34
& 44.90 & 58.50
& 11.63 & 26.82 \\
\midrule

\multirow{2}{*}{GFM-RAG} & Base
& 20.30 & 30.61
& 52.90 & 60.43
& 40.50 & 50.42
& 12.59 & 27.49 \\

& \ours
& 20.40 & 30.96
& \textbf{53.20} & 60.44
& 40.30 & 50.82
& 14.58 & 28.83 \\
\midrule

\multirow{2}{*}{HippoRAG2} & Base
& 27.90 & 39.95
& 51.20 & 57.53
& 53.80 & 68.03
& 17.59 & 32.72 \\

& \ours
& \textbf{29.30} & \textbf{41.28}
& 51.90 & 58.15
& \textbf{54.10} & \textbf{68.44}
& \textbf{18.40} & \textbf{33.81} \\

\bottomrule
\end{tabular}
}
\endgroup
\caption{Main results on multi-hop and long-document question answering benchmarks. Bold values indicate the best result in each metric column.}
\label{tab:main-results}
\end{table*}
We apply \ours to three GraphRAG frameworks that construct their graphs through LLM-based extraction: LightRAG~\citep{guo2024lightrag}, HippoRAG2~\citep{gutierrez2024hipporag}, and GFM-RAG~\citep{luo2026gfm}. For each, we compare the original base framework against its \ours-augmented variant. We additionally include LinearRAG~\citep{zhuang2025linearrag}, a non-LLM-based baseline, for broader context; \ours is not applied to LinearRAG, as its graph construction does not rely on LLM extraction.
\paragraph{Base Models}

We adopt Qwen3-32B~\citep{yang2025qwen3} as the backbone LLM; its 32,768-token context window provides sufficient capacity for the long evidence passages retrieved across our benchmarks. BGE-M3~\citep{chen2024bge} serves as the embedding model, and BAAI-bge-reranker-v2-m3 is used as the reranker where applicable. We set $k{=}5$ for top-$k$ retrieval across all frameworks.

\paragraph{Metrics}
We evaluate using Exact Match (EM) and F1 scores, which are widely used metrics in QA research~\citep{gutierrez2024hipporag, guo2024lightrag, luo2026gfm}.

\subsection{Experimental Results}



Table~\ref{tab:main-results} shows that \ours improves performance across all three base frameworks, showing that the proposed graph augmentation is framework-agnostic. 
Gains are most pronounced on MuSiQue and LiteraryQA,
where cross-chunk evidence plays a more prominent role. In particular, HippoRAG2+\ours achieves F1 improvements of +1.33 and +1.09 on these two benchmarks, respectively. Among all the results, LinearRAG achieves the highest F1 score (65.16) on 2WikiMultiHopQA despite using no LLM extraction, but falls behind HippoRAG2+\ours, which achieves the overall best performance, on the remaining three benchmarks.

\begin{figure*}[t]
\centering
\begin{minipage}[t]{0.49\textwidth}
\centering
\begin{tikzpicture}
\begin{axis}[
    ybar stacked,
    bar width=5pt,
    width=\linewidth,
    height=5.0cm,
    ylabel={False-to-True Cases (EM)},
    xtick={1,2,3,4,5,6,7,8,9,10,11,12},
    xticklabels={M, 2W, H, LQ, M, 2W, H, LQ, M, 2W, H, LQ},
    xticklabel style={font=\scriptsize},
    legend style={
        at={(0.5,1.13)},
        anchor=south,
        legend columns=2,
        font=\scriptsize,
        cells={anchor=west},
        draw=none,
        /tikz/every even column/.append style={column sep=4pt},
    },
    ymin=0,
    ymax=61,
    ymajorgrids=true,
    grid style={dashed, gray!30},
    tick style={draw=none},
    enlarge x limits=0.04,
    clip=false,
]
\addplot[fill=blue!35, draw=none]
    coordinates {
        (1,3) (2,5) (3,0) (4,3)
        (5,0) (6,0) (7,0) (8,0)
        (9,5) (10,3) (11,4) (12,5)
    };
\addplot[fill=orange!55, draw=none]
    coordinates {
        (1,11) (2,6) (3,3) (4,16)
        (5,2) (6,9) (7,1) (8,18)
        (9,12) (10,22) (11,18) (12,46)
    };
\addplot[fill=teal!55, draw=none]
    coordinates {
        (1,11) (2,5) (3,3) (4,0)
        (5,1) (6,2) (7,2) (8,0)
        (9,6) (10,14) (11,12) (12,0)
    };
\addplot[fill=purple!55, draw=none]
    coordinates {
        (1,0) (2,0) (3,0) (4,0)
        (5,2) (6,10) (7,9) (8,18)
        (9,0) (10,0) (11,0) (12,0)
    };
\addplot[fill=gray!40, draw=none]
    coordinates {
        (1,0) (2,0) (3,0) (4,0)
        (5,12) (6,4) (7,8) (8,5)
        (9,7) (10,6) (11,13) (12,3)
    };
\legend{Order-only, Set-only, New gold in top-$k$, Hidden facts in QA, Other F2T}
\draw[dashed, gray!45] (axis cs:4.5,0) -- (axis cs:4.5,61);
\draw[dashed, gray!45] (axis cs:8.5,0) -- (axis cs:8.5,61);
\node[anchor=north, font=\scriptsize] at (axis cs:2.5,-6) {HippoRAG2};
\node[anchor=north, font=\scriptsize] at (axis cs:6.5,-6) {LightRAG};
\node[anchor=north, font=\scriptsize] at (axis cs:10.5,-6) {GFM-RAG};
\node[above, font=\scriptsize\bfseries] at (axis cs:1,25) {25};
\node[above, font=\scriptsize\bfseries] at (axis cs:2,16) {16};
\node[above, font=\scriptsize\bfseries] at (axis cs:3,6) {6};
\node[above, font=\scriptsize\bfseries] at (axis cs:4,19) {19};
\node[above, font=\scriptsize\bfseries] at (axis cs:5,17) {17};
\node[above, font=\scriptsize\bfseries] at (axis cs:6,25) {25};
\node[above, font=\scriptsize\bfseries] at (axis cs:7,20) {20};
\node[above, font=\scriptsize\bfseries] at (axis cs:8,41) {41};
\node[above, font=\scriptsize\bfseries] at (axis cs:9,30) {30};
\node[above, font=\scriptsize\bfseries] at (axis cs:10,45) {45};
\node[above, font=\scriptsize\bfseries] at (axis cs:11,47) {47};
\node[above, font=\scriptsize\bfseries] at (axis cs:12,54) {54};
\end{axis}
\end{tikzpicture}
\caption{False-to-true EM case breakdown across three GraphRAG frameworks. Bars are stacked by retrieval effect; M (MuSiQue), 2W (2Wiki), H (HotpotQA), and LQ (LiteraryQA) denote the corresponding datasets.}
\label{fig:f2t-breakdown-all-frameworks}
\end{minipage}
\hfill
\begin{minipage}[t]{0.49\textwidth}
\centering

\begin{tikzpicture}
\begin{axis}[
    ybar,
    bar width=4pt,
    width=0.96\linewidth,
    height=5.7cm,
    ylabel={Cases (EM)},
    xtick={1,2,3,4,5,6,7,8,9,10,11,12,13,14,15,16},
    xticklabels={M, 2W, H, LQ, M, 2W, H, LQ, M, 2W, H, LQ, M, 2W, H, LQ},
    xticklabel style={font=\scriptsize},
    legend style={
        at={(0.5,1.12)},
        anchor=south,
        legend columns=2,
        font=\scriptsize,
        cells={anchor=west},
        draw=none,
        /tikz/every even column/.append style={column sep=5pt},
    },
    ymin=-48,
    ymax=56,
    ymajorgrids=true,
    grid style={dashed, gray!30},
    tick style={draw=none},
    enlarge x limits=0.02,
    clip=false,
]
\addplot[fill=blue!45, draw=none, bar shift=0pt]
    coordinates {
        (1,25) (2,16) (3,6) (4,19)
        (5,3) (6,11) (7,3) (8,18)
        (9,4) (10,21) (11,12) (12,34)
        (13,23) (14,39) (15,34) (16,51)
    };
\addplot[fill=red!45, draw=none, bar shift=0pt]
    coordinates {
        (1,-11) (2,-9) (3,-2) (4,-8)
        (5,-1) (6,-7) (7,-5) (8,-20)
        (9,-2) (10,-13) (11,-16) (12,-24)
        (13,-22) (14,-34) (15,-33) (16,-22)
    };
\legend{False-to-True, True-to-False}
\draw[black!45] (axis cs:0.5,0) -- (axis cs:16.5,0);
\draw[dashed, gray!45] (axis cs:4.5,-48) -- (axis cs:4.5,56);
\draw[dashed, gray!45] (axis cs:8.5,-48) -- (axis cs:8.5,56);
\draw[dashed, gray!45] (axis cs:12.5,-48) -- (axis cs:12.5,56);
\node[align=center, font=\tiny, inner sep=0pt] at (axis cs:2.5,-41) {HippoRAG2\\chunk};
\node[align=center, font=\tiny, inner sep=0pt] at (axis cs:6.5,-41) {LightRAG\\chunk};
\node[align=center, font=\tiny, inner sep=0pt] at (axis cs:10.5,-41) {LightRAG\\facts};
\node[align=center, font=\tiny, inner sep=0pt] at (axis cs:14.5,-41) {GFM-RAG\\chunk};
\end{axis}
\end{tikzpicture}
\caption{Bidirectional EM effect of retrieval changes. Upward bars show F2T cases, while downward bars show T2F cases, grouped by framework and change source.}
\label{fig:chunk-hidden-f2t-t2f}
\end{minipage}
\end{figure*}

\paragraph{Fine-grained Retrieval Analysis.}

We further examine the fine-grained effects of \ours on downstream retrieval and QA through false-to-true (F2T) cases, where the base framework fails but the \ours-augmented variant succeeds. As shown in Figure~\ref{fig:f2t-breakdown-all-frameworks}, these gains mainly arise from two retrieval effects. \emph{Chunk order change} keeps the retrieved chunk set unchanged but ranks more useful evidence higher, increasing its LLM attention. \emph{Chunk set change} alters the retrieved evidence by introducing either gold chunks into the top-$k$ or non-gold bridging chunks that still support the reasoning chain. In LightRAG, retrieved facts are directly inserted into the QA prompt, so \ours-extracted hidden facts can also provide compact relational evidence, accounting for many F2T improvements.
To assess whether these changes also produce harmful effects, we jointly examine F2T and true-to-false (T2F) cases caused by chunk changes and by hidden facts in the QA context. As shown in Figure~\ref{fig:chunk-hidden-f2t-t2f}, F2T counts consistently and substantially exceed the corresponding T2F counts across all frameworks and benchmarks. This confirms that the retrieval changes introduced by \ours are predominantly beneficial, primarily surfacing missing cross-chunk evidence rather than introducing misleading retrievals.

\subsection{Case Study}
We present representative cases illustrating \ours{}'s advantages in retrieval and downstream QA. We focus on a cross-chunk locative case where \ours recovers a relation unavailable from chunk-local triples alone, with additional cases in Appendix~\ref{sec:case-study}.
\paragraph{Recovering a cross-chunk locative relation.}
In \textit{Voodoo Planet}, the question asks: \textit{``Where does the confrontation between the Medic and Lumbrilo take place?''} The gold answer is \textit{in a deadly swamp}. The base framework answers \textit{``on a camping ground''}, whereas \ours answers that the confrontation occurs in the Khatkan poachers' camp within a swamp. The base graph contains locally correct triples such as \textit{(Tau, approached, Lumbrilo)}, \textit{(Tau, defeated, Lumbrilo)}, \textit{(Lumbrilo, practices, new magic)}, and \textit{(Voodoo Planet, features, showdown between Lumbrilo and Asaki)}. However, these triples only indicate that Tau and Lumbrilo are involved in a conflict; even when connected through multi-hop reasoning, these triples trace an event chain centering on Tau and Lumbrilo's conflict without specifying where the confrontation takes place. The base framework consequently forms an answer from incomplete context and misidentifies the location.

\ours extracts cross-chunk locative facts that directly connect the confrontation to the missing location, including \textit{(khatkan swamp, location of battle, lumbrilo vs tau and dane)} and \textit{(khatkan swamp, location of, final confrontation with lumbrilo)}. These facts are supported by the chunk that describes the swamp and camp together with the chunks that describe the battle itself. By adding these cross-chunk locative links to the graph index, \ours causes retrieval to surface the relevant swamp and confrontation passages:

\begin{quote}
\small
``The shadow thing in the swamp moved \ldots{} `I'd rather get at the source.' There was a grim note in the Medic's reply. `And to do that I want to look at their camp.'''
\end{quote}

This case illustrates the core failure mode targeted by \ours: the base graph contains true local event triples, but the cross-chunk relation needed to answer the question is absent from the index.

\subsection{Validation of Augmented Triple Quality}
\label{sec:deepseek-judge-validation}

\begin{center}
{\scriptsize
\setlength{\tabcolsep}{2pt}
\begin{tabular*}{\columnwidth}{@{\extracolsep{\fill}}lccc@{}}
\toprule
\textbf{Criterion} &
\textbf{DeepSeek Rate} &
\textbf{Human--LLM F1} &
\textbf{Cohen's $\kappa$} \\
\midrule
Factual correctness & 84.35 & 0.940 & 0.765 \\
Chunk ID correctness & 81.54 & 0.899 & 0.781 \\
\bottomrule
\end{tabular*}
}
\captionof{table}{Validation results for the DeepSeek judge on generated triples. DeepSeek correct rates are computed over parseable judge outputs, while Human--LLM F1 and Cohen's $\kappa$ are computed on the 100 manually annotated agreement samples.}
\label{tab:deepseek-judge-validation}
\end{center}

To validate the quality of triples produced by \ours, we randomly sample 1,000 augmented triples from each of the four datasets under the HippoRAG2+\ours setting and evaluate them using DeepSeek-V4-Pro as an automatic judge. Each triple is assessed on two criteria: factual correctness (whether the relation is accurate given the source document) and chunk citation accuracy (whether the cited chunk indeed supports the triple). To verify the reliability of this automatic evaluation, we further randomly sample 100 triples for human annotation on the same two criteria. We compute Human--LLM F1 and Cohen's $\kappa$ between human and automatic judgments to measure inter-rater agreement. Table~\ref{tab:deepseek-judge-validation} reports the results. The DeepSeek judge assigns high correct rates to both factual correctness (84.35\%) and chunk citation accuracy (81.54\%), indicating that the augmented triples are predominantly accurate and grounded in the cited evidence. The strong Human--LLM F1 scores (0.940 and 0.899) and substantial Cohen's $\kappa$ values (0.765 and 0.781) further confirm that the automatic evaluation closely mirrors human judgment, validating both the quality of \ours-extracted triples and the reliability of the automatic judge.

\subsection{Ablation Study}

\begin{table}[!htbp]
\centering
\small
\setlength{\tabcolsep}{0pt}
\begin{tabular*}{\columnwidth}{@{\extracolsep{\fill}}llcc@{}}
\toprule
\textbf{Dataset} & \textbf{Setting} & \textbf{EM} & \textbf{F1} \\
\midrule
\multirow{2}{*}{MuSiQue}
& \ours & 29.30 & 41.28 \\
& w/o GNN & 28.40 & 39.95 \\
\midrule
\multirow{2}{*}{2Wiki}
& \ours & 51.90 & 58.15 \\
& w/o GNN & 50.90 & 57.52 \\
\midrule
\multirow{2}{*}{HotpotQA}
& \ours & 54.10 & 68.44 \\
& w/o GNN & 53.30 & 67.66 \\
\midrule
\multirow{2}{*}{LiteraryQA}
& \ours & 18.40 & 33.81 \\
& w/o GNN & 18.19 & 33.17 \\
\bottomrule
\end{tabular*}
\caption{Ablation study on the GNN scorer in HippoRAG2. The w/o GNN setting replaces \ours's scorer-selected subgraphs with the same number of randomly selected subgraphs for LLM completion.}
\label{tab:ablation-random-subgraph}
\end{table}

To assess the GNN scorer, we replace it with random subgraph selection under the same LLM completion budget, isolating the effect of subgraph selection quality. As shown in Table~\ref{tab:ablation-random-subgraph}, this replacement degrades performance on all four benchmarks, especially MuSiQue (F1: $-$1.33; EM: $-$0.90) and 2WikiMultiHopQA (EM: $-$1.00). These results show that topology-guided scoring improves subgraph selection and is effective in \ours.

\section{Conclusion}

We present \ours, a GNN-guided graph completion method that addresses the chunk-local extraction bottleneck in GraphRAG. \ours operates through three coordinated modules: a self-supervised graph corruption module that derives training supervision directly from the base graph without manual annotation; a topology-aware GNN scoring module that identifies structurally incomplete subgraphs; and a GNN-guided LLM completion module that produces evidence-grounded triples for the selected regions. Applied to three LLM-based GraphRAG frameworks across four QA benchmarks, \ours consistently improves performance. By restricting LLM completion to a targeted subset of high-missingness subgraphs, \ours produces a persistent augmented graph that can be reused by any downstream retrieval pipeline without further recomputation.

\clearpage
\section*{Limitations}

\ours introduces additional indexing cost primarily through prompt tokens, as each LLM completion call is supplied with a local subgraph, multiple evidence passages, and known triples. As discussed in Appendix~\ref{sec:cost-sensitivity}, the GNN scorer limits LLM completion to a small subset of selected subgraphs, keeping the additional call count and wall-clock time to modest fractions of the base pipeline overhead. The overall additional cost is therefore controllable through the GNN scoring threshold and LLM budget parameters. Nevertheless, the prompt overhead per call is substantial by design, and future work could explore more efficient context compression or selective evidence retrieval strategies to further reduce this cost without sacrificing completion quality.

\section*{Ethical considerations}

\ours augments GraphRAG indices with LLM-generated triples; despite our citation-grounding filter and the validation rates above 81\% reported in Section~\ref{sec:deepseek-judge-validation}, residual extraction errors may propagate to downstream QA, and practitioners should verify the augmented triples in high-stakes settings. All four English QA benchmarks (MusiQue, 2WikiMultiHopQA, HotpotQA, LiteraryQA) and all models used (Qwen3-32B, BGE-M3, BAAI-bge-reranker-v2-m3, DeepSeek-V4-Pro) are publicly released for research and are used in a manner consistent with their original intended purposes under their respective licenses; the underlying corpora are drawn from Wikipedia and public-domain literary works and contain no private personal information, so no anonymization or participant consent was required. Dataset statistics, train/test splits, and graph statistics are reported in Table~\ref{tab:datasets}; the model configurations, GNN scorer ($\sim$$10^6$ parameters), A800-based infrastructure, and full indexing-time compute budget (LLM calls, prompt/completion tokens, and wall-clock hours) are reported in Appendix~\ref{sec:cost-sensitivity} and Table~\ref{tab:cost-efficiency}, with existing packages used at default settings. The 100 manually annotated triples in Section~\ref{sec:deepseek-judge-validation} were labeled internally by the authors on two binary criteria, so no external recruitment, payment, consent procedure, or ethics-board review was applicable. We release the source code under a permissive open-source license, and AI assistants were used only for editorial polishing and routine code suggestions, with all scientific claims authored and verified by the human authors.

\bibliography{main}

\clearpage
\appendix

\onecolumn
\section{Prompt}
\label{sec:prompt}
\begin{center}
\includegraphics[width=\textwidth,height=0.86\textheight,keepaspectratio]{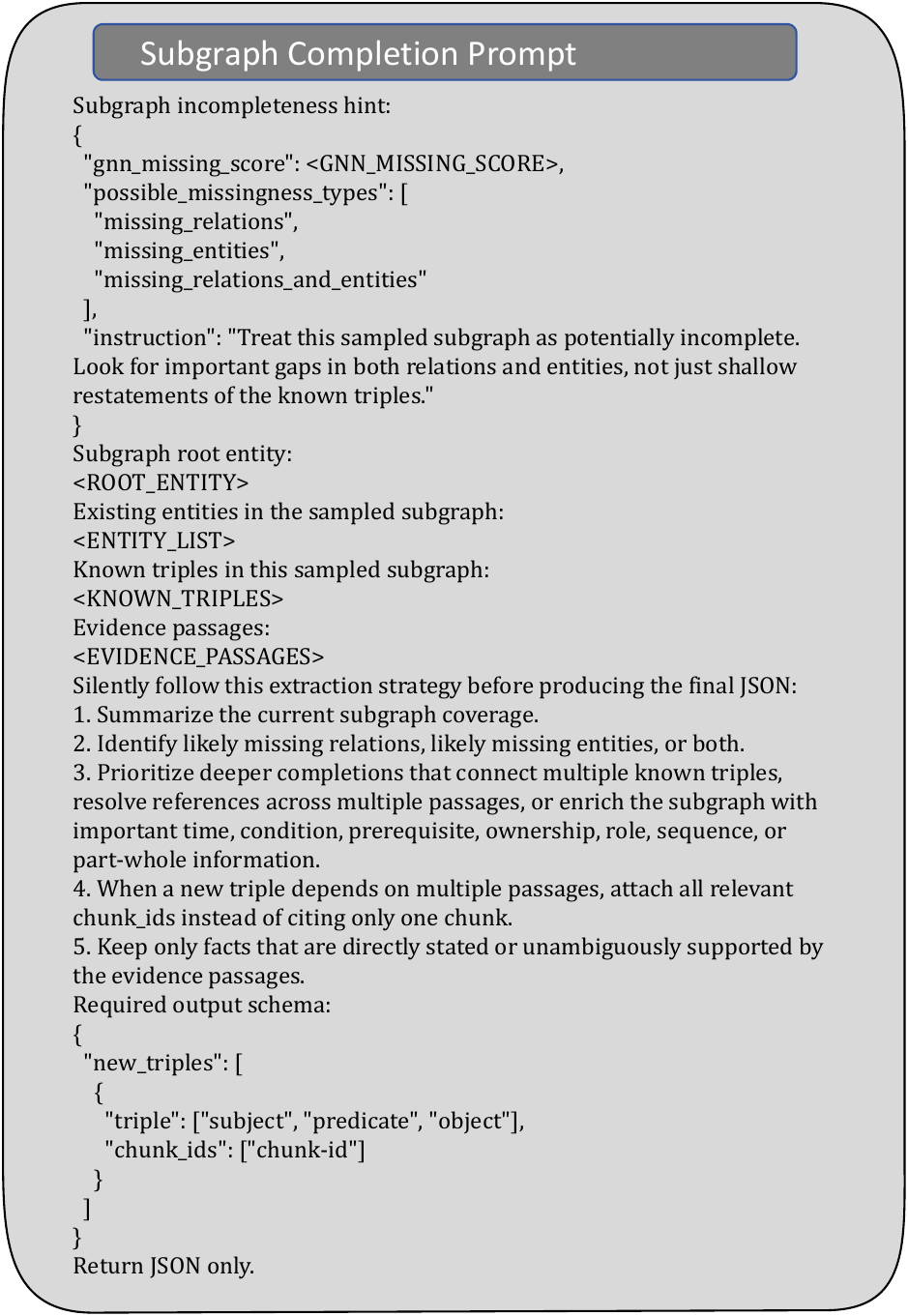}
\captionof{figure}{Prompt used by \ours for GNN-guided LLM completion. The selected subgraph, known triples, candidate entities, and evidence chunks are provided to the LLM, which is instructed to return only evidence-supported triples and entities.}
\label{fig:completion-prompt}
\end{center}

\clearpage
\twocolumn
\section{Case Study}
\label{sec:case-study}

We provide additional representative examples where \ours improves downstream QA performance. Each case illustrates a distinct failure mode of the base framework---misidentified entities, absent relational links, or insufficiently ranked evidence---and shows how cross-chunk augmentation addresses it by enriching the graph index before retrieval.

\paragraph{Recovering the correct object in long-document QA.}
In \textit{Immensee}, the question asks: \textit{``What did Erich give Elisabeth?''} The gold answer is \textit{a bird} or \textit{a new bird}. The base framework answers \textit{``The gilded birdcage''}, while \ours answers \textit{``A canary''}. The base retrieval returns chunks 0025, 0026, 0013, 0021, and 0027, which contain related narrative context but do not directly identify the gift. \ours instead generates hidden relational facts around the same narrative neighborhood, such as \textit{(Eric, is, friend of Elisabeth)} and \textit{(Elisabeth, accepted proposal from, Eric)}. These triples do not directly state the answer, but they make the Eric--Elisabeth relation more explicit in the graph and change retrieval toward the passage where Eric's gift is described. As a result, \ours replaces chunk 0027 with chunk 0014, which already contains the base-extracted fact \textit{(Eric, sent canary to, Elisabeth)} and the decisive textual evidence:

\begin{quote}
\small
``Elisabeth was standing by the window and sticking some fresh chick-weed in a gilded birdcage which he had not seen in the place before. In the cage was a canary \ldots{} `Your friend Eric sent it this noon from his estate as a present for Elisabeth.'''
\end{quote}

This example illustrates a common long-document failure mode in which the retrieved context contains a salient object near the answer but not the answer-defining relation. The birdcage is strongly associated with the scene and with Elisabeth, so it is a plausible distractor for the base framework. However, knowing that Elisabeth is near a birdcage, or that Erich is associated with Elisabeth elsewhere in the story, is not sufficient to infer what Erich gave her. The retrieval improvement therefore comes from a two-step effect: \ours first strengthens the graph neighborhood connecting the giver and recipient, and this modified graph then surfaces chunk 0014, where the canary is explicitly linked to Eric as the sender and Elisabeth as the recipient of the present. By shifting retrieval to this chunk, \ours prevents the LLM from answering with a nearby container and instead grounds the answer in the entity that actually satisfies the gift relation.

\paragraph{Using hidden facts to bridge retrieval.}
On MuSiQue, the question asks: \textit{``What was the date that Pam got married in The Office?''} The gold answer is \textit{October 8, 2009}. The base framework answers that the date is not mentioned in the provided information, whereas \ours answers \textit{``October 8, 2009.''} The base retrieval mainly returns general \textit{Pam Beesly} and \textit{The Office} passages. These passages state that Jim and Pam marry in the episode \textit{Niagara}, but they do not directly surface the episode date:

\begin{quote}
\small
``Jim and Pam marry early in the season, at Niagara Falls, during the highly anticipated, hour long episode, `Niagara'.''
\end{quote}

\ours extracts hidden facts that connect the marriage event to the episode, including \textit{(niagara the office, features, pam beesly and jim halpert s marriage)} and \textit{(the episode, features, jim and pam s marriage at niagara falls)}. These hidden facts add a more direct graph link between the marriage event and the episode. As a result, retrieval includes the \textit{Niagara} episode page, which contains the exact answer:

\begin{quote}
\small
``Niagara'' \textit{The Office} episode \ldots{} Original air date October 8, 2009.
\end{quote}

The base triples can connect \textit{Pam}, \textit{Jim}, and the wedding event, but they do not reliably bridge that event to the episode page that stores the air date. This case shows that \ours does not merely add more triples to the index. The completed graph can change the starting points of graph retrieval, allowing downstream QA to reach evidence that the base graph fails to use.

\paragraph{Correcting an entity through better evidence retrieval.}
In \textit{Youth}, the question asks: \textit{``What is the name of the steamer that agrees to tow the crew?''} The gold answer is \textit{Sommerville}. The base framework answers \textit{``The Judea''}, confusing the damaged ship with the steamer that tows it. \ours answers \textit{``The Sommerville.''} The base retrieval surfaces narrative passages from \textit{Youth}, including chunks 0001, 0006, 0016, 0012, and 0027. These passages mention the ship \textit{Judea} and earlier accidents but do not identify the towing steamer. \ours instead retrieves chunks 0020 and 0021, where the towing event is explicitly described:

\begin{quote}
\small
``When our skipper came back we learned that the steamer was the Sommerville, Captain Nash \ldots{} and that the agreement was she should tow us to Anjer or Batavia, if possible \ldots{} At noon the steamer began to tow.''
\end{quote}

The following chunk continues the same event:

\begin{quote}
\small
``The speed of the towing had fanned the smoldering destruction \ldots{} `We had better stop this towing'.''
\end{quote}

The base answer is plausible precisely because \textit{Judea} is the central vessel throughout the narrative. The question, however, asks for the steamer that performs the towing, not the ship being towed. The base evidence associates a prominent entity with the question but does not resolve the specific relational role it asks about. \ours retrieves the event-local passages that explicitly distinguish the vessel doing the towing from the one being towed.

\paragraph{Improving evidence ranking with a hidden date relation.}
Another MuSiQue example asks: \textit{``What date saw the writing of the song where the devil went down to the state where WDXQ is located?''} The gold answer is \textit{May 21, 1979}. The base framework answers \textit{``1978''}, while \ours answers \textit{``May 21, 1979.''} Both frameworks retrieve passages about \textit{The Devil Went Down to Georgia} and WDXQ, but \ours moves the most useful evidence from rank 2 to rank 1. The retrieved WDXQ passage establishes the state:

\begin{quote}
\small
``WDXQ (1440 AM) is a radio station licensed to Cochran, Georgia, United States.''
\end{quote}

The decisive song passage states:

\begin{quote}
\small
``The Devil Went Down to Georgia'' \ldots{} Released May 21, 1979 \ldots{} Songwriter(s) Charlie Daniels, Tom Crain, ``Taz'' DiGregorio, Fred Edwards, Charles Hayward, James W. Marshall.
\end{quote}

\ours introduces the hidden fact \textit{(the devil went down to georgia, released in, may 21 1979)}, which links the song directly to its release date in the graph index. The base evidence can connect WDXQ to Georgia and the song to adjacent release context, but without the precise date relation it is susceptible to anchoring on an approximate nearby year such as \textit{1978}. This example highlights a benefit distinct from the preceding cases: \ours need not always introduce an entirely new document into retrieval. By anchoring retrieval to this hidden date triple, it promotes already-relevant evidence to a higher rank, enabling the LLM to ground its answer in the passage that contains the exact date.

\section{Cost and Performance Across Parameter Settings}
\label{sec:cost-sensitivity}

\ours exposes an LLM completion budget parameter that controls how many candidate subgraphs receive LLM completion calls during indexing. Adjusting this budget allows practitioners to navigate the trade-off between augmentation coverage and indexing overhead. Table~\ref{tab:cost-efficiency} reports the overhead introduced by \ours at the default budget setting on LiteraryQA, where \ours adds 1,483 LLM calls, 17.66M prompt tokens, and 5.7 hours of indexing time on top of the base pipeline---approximately 8.2\% and 15.3\% of the base's calls and wall-clock time. Each prompt is substantially longer than in the base pipeline because it incorporates a local subgraph, multiple evidence passages spanning different chunks, known triples, and entity context; completion overhead is comparatively small, as the model outputs only evidence-supported triples.

\begin{figure}[!htbp]
\centering
\begin{tikzpicture}
\begin{groupplot}[
    group style={
        group size=1 by 3,
        vertical sep=1.05cm
    },
    width=0.90\columnwidth,
    height=0.36\columnwidth,
    xmin=40,
    xmax=210,
    xtick={50,100,150,200},
    grid=major,
    grid style={gray!15},
    tick label style={font=\scriptsize},
    label style={font=\scriptsize},
    title style={font=\footnotesize\bfseries},
]
\nextgroupplot[
    title={(a) EM},
    ylabel={EM},
    ymin=18.1,
    ymax=18.9,
    ytick={18.2,18.4,18.6,18.8},
]
\addplot+[blue!70!black, thick, mark=*, mark size=1.8pt]
coordinates {
    (50,18.26)
    (100,18.40)
    (150,18.48)
    (200,18.78)
};
\nextgroupplot[
    title={(b) F1},
    ylabel={F1},
    ymin=33.55,
    ymax=34.10,
    ytick={33.6,33.8,34.0},
]
\addplot+[red!70!black, thick, mark=square*, mark size=1.8pt]
coordinates {
    (50,33.67)
    (100,33.81)
    (150,33.80)
    (200,34.00)
};
\nextgroupplot[
    title={(c) Completion Tokens},
    xlabel={Budget},
    ylabel={Tokens ($\times 10^6$)},
    ymin=2.0,
    ymax=6.5,
    ytick={2,3,4,5,6},
]
\addplot+[teal!70!black, thick, mark=triangle*, mark size=2.0pt]
coordinates {
    (50,2.58)
    (100,2.63)
    (150,5.38)
    (200,5.99)
};
\end{groupplot}
\end{tikzpicture}
\caption{Budget sensitivity on LiteraryQA. Panels report EM, F1, and completion-token overhead as the LLM completion budget changes.}
\label{fig:budget-trends}
\end{figure}
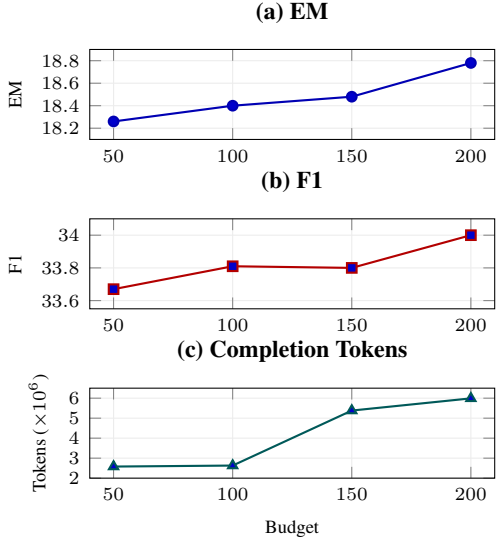

Figure~\ref{fig:budget-trends} shows how EM and F1 respond as the budget is varied from 50 to 200 on LiteraryQA. Both metrics improve as the budget increases while completion-token cost scales proportionally. A budget of 50 already yields measurable gains over the base pipeline, showing that useful augmentation is achievable at modest additional cost; increasing the budget to 200 provides further, consistent improvement. This controllable mechanism allows users to select an operating point that matches their latency and cost requirements.

\begin{table}[!htbp]
\centering
\small
\setlength{\tabcolsep}{3pt}
\begin{tabular}{@{}lrrrr@{}}
\toprule
\textbf{Setting} & \textbf{Prompt} & \textbf{Completion} & \textbf{LLM} & \textbf{Time} \\
 & \textbf{($\times 10^6$)} & \textbf{($\times 10^6$)} & \textbf{Calls} & \textbf{(h)} \\
\midrule
Base & 17.83 & 23.02 & 17,990 & 37.2 \\
\ours & 17.66 & 2.63 & 1,483 & 5.7 \\
\bottomrule
\end{tabular}
\caption{Cost comparison between the HippoRAG2 base pipeline and \ours. The \ours row reports the additional overhead introduced by \ours on top of the base pipeline. Prompt and completion token counts are reported in millions of tokens ($\times 10^6$).}
\label{tab:cost-efficiency}
\end{table}


\FloatBarrier

\section{Hyperparameter Sensitivity Analysis}
\label{sec:param-sensitivity}

We further analyze the sensitivity of \ours{}'s performance to key hyperparameters. Specifically, we consider the LLM completion budget ($B$), the fact-edge mask ratio ($\rho_e$), and the entity-node deletion ratio ($\rho_v$), varying one hyperparameter at a time while fixing the others. The results show that \ours is robust against these hyperparameters across a practical range of values.

\begin{table}[!htbp]
\centering
\footnotesize
\setlength{\tabcolsep}{3pt}
\begin{tabular*}{\columnwidth}{@{\extracolsep{\fill}}lrrrrr@{}}
\toprule
\begin{tabular}[c]{@{}c@{}}\textbf{Fact}\\\textbf{Mask}\end{tabular} &
\textbf{EM} & \textbf{F1} & \textbf{Calls} &
\begin{tabular}[c]{@{}c@{}}\textbf{Prompt}\\\textbf{tokens}\end{tabular} &
\begin{tabular}[c]{@{}c@{}}\textbf{Completion}\\\textbf{tokens}\end{tabular} \\
 & & & & \multicolumn{2}{c}{\textbf{($\times 10^6$)}} \\
\midrule
0.10 & 17.60 & 32.96 & 2,313 & 27.79 & 4.02 \\
0.15 & 17.60 & 32.99 & 1,577 & 18.70 & 2.84 \\
0.20 & 18.40 & 33.81 & 1,483 & 17.66 & 2.63 \\
0.25 & 17.45 & 33.41 & 1,031 & 11.97 & 1.83 \\
\bottomrule
\end{tabular*}
\caption{Hyperparameter sensitivity analysis for the fact-mask ratio on (HippoRAG2+\ours, LiteraryQA). Prompt and completion token counts are reported in millions of tokens ($\times 10^6$).}
\label{tab:fact-mask-sensitivity}
\end{table}

The default ratio $\rho_e{=}0.20$ achieves the best performance (EM 18.40, F1 33.81). Lower ratios (0.10, 0.15) produce corruption signals that are too subtle, yielding a less discriminative scorer that passes more subgraphs to the LLM (up to 2,313 calls) without improving QA performance. A ratio of 0.25 over-corrupts the training examples, removing so many edges that the surviving subgraph structure is too sparse to provide useful training signal, reducing both LLM calls and performance.

\begin{table}[!htbp]
\centering
\footnotesize
\setlength{\tabcolsep}{3pt}
\begin{tabular*}{\columnwidth}{@{\extracolsep{\fill}}lrrrrr@{}}
\toprule
\begin{tabular}[c]{@{}c@{}}\textbf{Entity}\\\textbf{Mask}\end{tabular} &
\textbf{EM} & \textbf{F1} & \textbf{Calls} &
\begin{tabular}[c]{@{}c@{}}\textbf{Prompt}\\\textbf{tokens}\end{tabular} &
\begin{tabular}[c]{@{}c@{}}\textbf{Completion}\\\textbf{tokens}\end{tabular} \\
 & & & & \multicolumn{2}{c}{\textbf{($\times 10^6$)}} \\
\midrule
0.06 & 17.53 & 33.02 & 1,309 & 12.91 & 1.83 \\
0.08 & 18.40 & 33.81 & 1,483 & 17.66 & 2.63 \\
0.10 & 17.89 & 33.44 & 1,433 & 16.83 & 2.57 \\
0.12 & 17.67 & 32.93 & 1,400 & 16.12 & 2.53 \\
\bottomrule
\end{tabular*}
\caption{Hyperparameter sensitivity analysis for the entity-mask ratio on (HippoRAG2+\ours, LiteraryQA). Prompt and completion token counts are reported in millions of tokens ($\times 10^6$).}
\label{tab:entity-mask-sensitivity}
\end{table}

The default ratio $\rho_v{=}0.08$ yields the best performance (EM 18.40, F1 33.81). A smaller ratio (0.06) produces insufficiently corrupted views, limiting the GNN's ability to learn entity-level absence patterns and resulting in fewer selected subgraphs (1,309). Larger ratios (0.10, 0.12) remove too many entities, stripping the training examples of the relational structure needed for discriminative learning. The relatively narrow range of performance differences across all four settings indicates that \ours is moderately robust to the entity-mask ratio.

\begin{table}[!htbp]
\centering
\footnotesize
\setlength{\tabcolsep}{3pt}
\begin{tabular*}{\columnwidth}{@{\extracolsep{\fill}}lrrrrrr@{}}
\toprule
\begin{tabular}[c]{@{}c@{}}\textbf{Score}\\\textbf{($\theta$)}\end{tabular} &
\textbf{EM} & \textbf{F1} & \textbf{LLM} & \textbf{Calls} &
\begin{tabular}[c]{@{}c@{}}\textbf{Prompt}\\\textbf{tokens}\end{tabular} &
\begin{tabular}[c]{@{}c@{}}\textbf{Completion}\\\textbf{tokens}\end{tabular} \\
 & & & & & \multicolumn{2}{c}{\textbf{($\times 10^6$)}} \\
\midrule
0.40 & 18.58 & 34.24 & 53.30 & 3,788 & 40.42 & 6.38 \\
0.45 & 18.43 & 34.02 & 53.04 & 2,452 & 32.48 & 5.22 \\
0.50 & 18.40 & 33.81 & 53.09 & 1,483 & 17.66 & 2.63 \\
0.55 & 17.92 & 33.54 & 51.59 &   760 &  8.57 & 1.36 \\
\bottomrule
\end{tabular*}
\caption{Hyperparameter sensitivity analysis for GNN score threshold $\theta$ on (HippoRAG2+\ours, LiteraryQA). Prompt and completion token counts are in millions ($\times 10^6$).}
\label{tab:score-threshold}
\end{table}

Lowering $\theta$ allows more subgraphs to pass the GNN filter, increasing both coverage and LLM cost. At $\theta{=}0.40$, \ours selects 3,788 subgraphs and achieves the best EM (18.58) and F1 (34.24), but at more than 2.5$\times$ the token cost of the default ($\theta{=}0.50$). At $\theta{=}0.55$, the call count drops to 760 and prompt tokens fall to 8.57M, but both EM and F1 also decline below the default. The default $\theta{=}0.50$ provides a balanced operating point; practitioners can shift $\theta$ to trade performance for cost as their indexing budget allows.

\FloatBarrier
\section{Human Preference Trial and Evaluation Details}
\label{sec:human-eval-details}

This appendix documents the protocol behind the human evaluation summarized in Section~\ref{sec:deepseek-judge-validation} of the main text, whose purpose is twofold: to estimate the absolute quality of \ours-augmented triples and to verify that the DeepSeek-V4-Pro judge produces decisions consistent with informed human raters.

\paragraph{Annotator profile.}
Three of the paper authors served as annotators. All three have prior research experience with knowledge graphs and retrieval-augmented generation, so they can interpret triple syntax and reason about whether a cited chunk supports a triple. No external annotators or crowdworkers were recruited, and no monetary compensation, recruitment platform, or demographic balancing was involved.

\paragraph{Sampling.}
From the pool of \ours-augmented triples produced by HippoRAG2+\ours on the four benchmarks, we drew a stratified random sample of 100 triples (25 per benchmark). Stratification ensures that domain-specific failure modes---encyclopedic factoid linkage in HotpotQA/2Wiki, multi-hop composition in MusiQue, and narrative coreference in LiteraryQA---are equally represented in the human-labeled set.

\paragraph{Rubric.}
Each item presented to an annotator includes the triple in \emph{(subject, relation, object)} form, the cited chunk identifier(s), and the full chunk text. Annotators returned a strict binary judgment on two independent criteria:
\begin{itemize}[noitemsep,topsep=2pt]
    \item \textbf{Factual correctness}---the triple expresses a relation that is true with respect to the cited evidence and the document context.
    \item \textbf{Chunk-citation accuracy}---at least one cited chunk contains direct textual evidence for the triple.
\end{itemize}
The instructions explicitly state that the two criteria are decided independently: a triple may be factually correct overall while its citation fails to provide direct support, and vice versa.

\paragraph{Procedure and quality control.}
Each triple was labeled independently by all three annotators without communication during labeling. The final binary label per criterion is the majority vote across the three annotators; disagreements were logged for post-hoc inspection. Annotators were instructed to rely only on the cited chunk text and to suppress external knowledge---a rule that is especially important for LiteraryQA, where annotators may recognize the source works.

\paragraph{Agreement with the automatic judge.}
Inter-rater agreement between the majority human label and the DeepSeek-V4-Pro decision is reported as Human--LLM F1 and Cohen's $\kappa$ in Table~\ref{tab:deepseek-judge-validation}. The $\kappa$ values of 0.765 (factual correctness) and 0.781 (chunk-citation accuracy) fall within the substantial-agreement range under common interpretation, justifying the use of the DeepSeek judge as a scalable proxy for the larger per-dataset evaluation of 1{,}000 triples.

\paragraph{Risks and consent.}
The annotation task involved only publicly available text (Wikipedia excerpts and public-domain literary passages) and posed no foreseeable risks to annotators. Because all annotation was internal to the research team and used only publicly released material, no external consent procedure or ethics-board review was applicable.

\end{document}